\documentclass{esannV2}
\usepackage{graphicx}
\usepackage[latin1]{inputenc}
\usepackage{amssymb,amsmath,array}
\usepackage{subfigure}
\usepackage{sidecap}

\begin{document}

\title{Importance of user inputs while using incremental learning to personalize human activity recognition models}

\author{Pekka Siirtola, Heli Koskim\"{a}ki and Juha R\"{o}ning
%
%
\vspace{.3cm}\\
Biomimetics and Intelligent Systems Group,\\ P.O. BOX 4500, FI-90014, University of Oulu, Oulu, Finland\\
\{pekka.siirtola, heli.koskimaki, juha.roning\}@oulu.fi
}

\maketitle


%
%
%
\hyphenation{manu-scripts manu-script ext-re-mums user-in-de-pen-dent user-de-pen-dent in-de-pen-dent mo-del in-de-pen-dent}

\begin{abstract}
In this study, importance of user inputs is studied in the context of personalizing human activity recognition models using incremental learning. Inertial sensor data from three body positions are used, and the classification is based on Learn++ ensemble method. Three different approaches to update models are compared: non-supervised, semi-supervised and supervised. Non-supervised approach relies fully on predicted labels, supervised fully on user labeled data, and the proposed method for semi-supervised learning, is a combination of these two. In fact, our experiments show that by relying on predicted labels with high confidence, and asking the user to label only uncertain observations (from 12\% to 26\% of the observations depending on the used base classifier), almost as low error rates can be achieved as by using supervised approach. In fact, the difference was less than 2\%-units. Moreover, unlike non-supervised approach, semi-supervised approach does not suffer from drastic concept drift, and thus, the error rate of the non-supervised approach is over 5\%-units higher than using semi-supervised approach.
\end{abstract}


%
\section{Problem statement and related work}
\label{problem}
\vspace{-0.5em}

This study focuses on human activity recognition based on inertial sensor data collected using smartphone sensors. One of the main challenges of the field is that people are different: they are unique for instance in terms of physical characteristics, health state or gender. Due to this, it is shown that a model that provides accurate results for one person, does not necessarily work accurately with somebody else's data. For instance, user-independent models are not accurate if they are trained with healthy study subjects and tested with subjects who have difficulties to move \cite{albert2012using}. Personal recognition models provide better recognition rates, but the challenge is that they normally require personal training data, and therefore, a personal data gathering session \cite{bulling}. 

Our previous study \cite{siirtola2018ESANN} presented for the first time how incremental learning can be used to personalize activity recognition models without a separate data gathering session. The advantage of incremental learning is that models can continuously learn from streaming data, and therefore, adapt to the user's personal moving style. In addition, learning can be done without model re-training, instead, models are \textit{updated} based on streaming data, and therefore, all the training data does not need to be kept stored. In the article, the idea was to base the recognition in the first place in a user-independent model, and use streaming data and predicted labels related to them to update the model, and therefore, personalize it. While the idea of learning without user-interruption is great from the user experience point-of-view, due to concept drift it can also lead to an unwanted end-result. Microsoft's self-learning Twitter chatbot Tay is a perfect example of what can happen if self-learning artificial intelligence applications learn wrong things. The idea of Tay was to mimic a 19-year-old girl and learn from interactions between other Twitter users. However, due to a drastic concept drift, it quickly learned how to be a racist and act like one because it leaned wrong things from online data \cite{tay}. 
This could have been avoided if, instead of self-learning, developers of Tay would have selected which interactions are used to update Tay's models and which not.

A different approach to personalize a human activity recognition model incrementally was presented in \cite{mannini2018}. The idea of the study was to avoid a drastic concept drift by using active learning: models were updated based on user labeled instances. Moreover, unlike in our previous study, models were not personalized using all the streaming data. Instead, the user was asked to label only uncertain observations, and model update was based on only these. When the method was tested using the leave-one-subject-out -method, the overall accuracy improved only slightly, however: from 88.6\% (without personalization) to 89.6\% (every other observation labeled by the user and used for personalization). However, it still improved the recognition accuracy of 40 subjects out of 53.

In this study, we combine ideas presented in \cite{siirtola2018ESANN} and \cite{mannini2018}. We use the whole streaming data to personalize classification models, as done in \cite{siirtola2018ESANN}. However, to avoid a drastic concept drift, we do not completely trust on predicted labels. Instead, we ask the user to label uncertain observations, as done in \cite{mannini2018}. This approach is compared to the approach presented in our previous study, and to the approach where the user labels every single instance from the streaming data. 
Moreover, the experiments done in this study are more comprehensive than the ones done in our previous study, now data from three body positions are studied while our previous study used data from one body position only.


\section{Experimental dataset}
\label{data}
\vspace{-0.5em}

In this study, a publicly open data set presented in \cite{Shoaib} was used in the experiments. It contains accelerometer, magnetometer and gyroscope data from seven physical activities (walking, sitting, standing, jogging, biking, walking upstairs and downstairs). The data were collected using a smartphone from ten study subjects using a sampling rate of 50Hz, and five body locations. In this study we selected to use accelerometer and gyroscope data from three positions: waist, wrist, and arm. Moreover, apparently one of the study subjects had a smartphone in different orientation than others making the data totally different to other subjects' data. Thus, only nine persons data were  used in the experiments.

In the study, features were extracted from windows of size 4.2 second, and 1.4 second slide was used. Used features were the ones that are commonly used in activity recognition studies. These include standard deviation, minimum, maximum, median, and different percentiles (10, 25, 75, and 90). Other used features were the sum of values above or below a given percentile (10, 25, 75, and 90), square sum of values above or below a given percentile (10, 25, 75, and 90), and number of crossings above or below a given percentile (10, 25, 75, and 90), and features from the frequency domain. These were extracted from raw accelerometer and gyroscope signals, as well as from magnitude signals and signals where two out of three accelerometer and gyroscope signals were square summed. Altogether 244 features were extracted.

\section{Methods for personalizing the recognition model}
\label{aim}
\vspace{-0.5em}

Learn++ algorithm \cite{polikar2001learn++} for incremental learning is an ensemble method, which can use any classifier as a base classifier.  In this study, three different base classifiers are compared: CART (classification and regression tree), LDA (linear discriminant analysis) and QDA (quadratic discriminant analysis). Learn++ processes incoming streaming data as chunks. For each chunk, a new group of weak base models are trained and combined to a group of previously trained base models as an ensemble model \cite{hammerchoosing}. Because of encouraging experiences obtained in our previous study, Learn++ is used also in this study.

There are three different ways to update models: non-supervised, semi-supervised, and supervised approach. Our previous article \cite{siirtola2018ESANN} used the non-supervised approach: only predicted labels were used to update the model. The semi-supervised approach uses both predicted labels and true labels for updating the model. Moreover, in the  supervised approach (a.k.a. active learning) each observation is labeled by the user, therefore, only true labels are used.

A new approach for semi-supervised learning is introduced in this study. The idea is that predicted labels with high priory and data related to them are used in the model update process as they are, but labels with a priori confidence below some threshold, $th$, are considered as uncertain, and they are labeled by the user before they are used in the model updating process. Moreover, as the studied activities are of long-term \cite{siirtola2011periodic} nature, it can be assumed that also the windows right before and after have the same class label as the window labeled by the user. In this study, we assume that if the label of window $w$ is $a$, then also windows $w-2, w-1, w+1$ and $w+2$ belong to class $a$. This way, one user input gives information to more than one window, improving the accuracy of the labels used to update models.

While non-supervised, semi-supervised, and supervised method use different approaches to label the data, otherwise, they are used to personalizing the human activity recognition models in the same way. These approaches are experimented using the leave-one-subject-out-method which is explained in \cite{siirtola2018ESANN}, and it can be divided into three steps. To personalize a model for subject $s_n$, Step 1 is to train a Learn++ based user-independent recognition model by randomly sampling data from subjects $s_1, ..., s_{n-1}$, and select the best features for the model using SFS (sequential forward selection). A new base model is then trained based on them and it is added to the ensemble of models. The ensemble model is then tested using the test data set. Step 1 is repeated $n$ times, in this case $n=3$. 
In Step 2, the personalization of the model starts by extracting features from the first chunk of subject $s_n$'s data. Depending on the used personalization approach, this data is labeled using the ensemble model, user or combination of these two depending on which approach is used. The data chunks used to personalize the recognition tend to be small, and therefore, the inner-variance of data chunks is small as well, leading easily to over-fitted models. To avoid this, variance and the amount of the data are increased using the noise injection method presented in \cite{cidm2016}. Next, the data used in model updating is selected based on random sampling, and best features are selected using SFS. These are then used to train a new base model, which is added to the ensemble. Step 2 is repeated $n$ times. Step 3 is to be done as in Step 2 but with a new data chunk. Therefore, in this study, in the end the Learn++ conducts of $3 \cdot n = 9$ base models. Equal weight is given to each base model.

\begin{figure}[t]
\begin{center}
 		\subfigure[Arm / LDA]{\label{f1}\includegraphics[width=3.75cm]{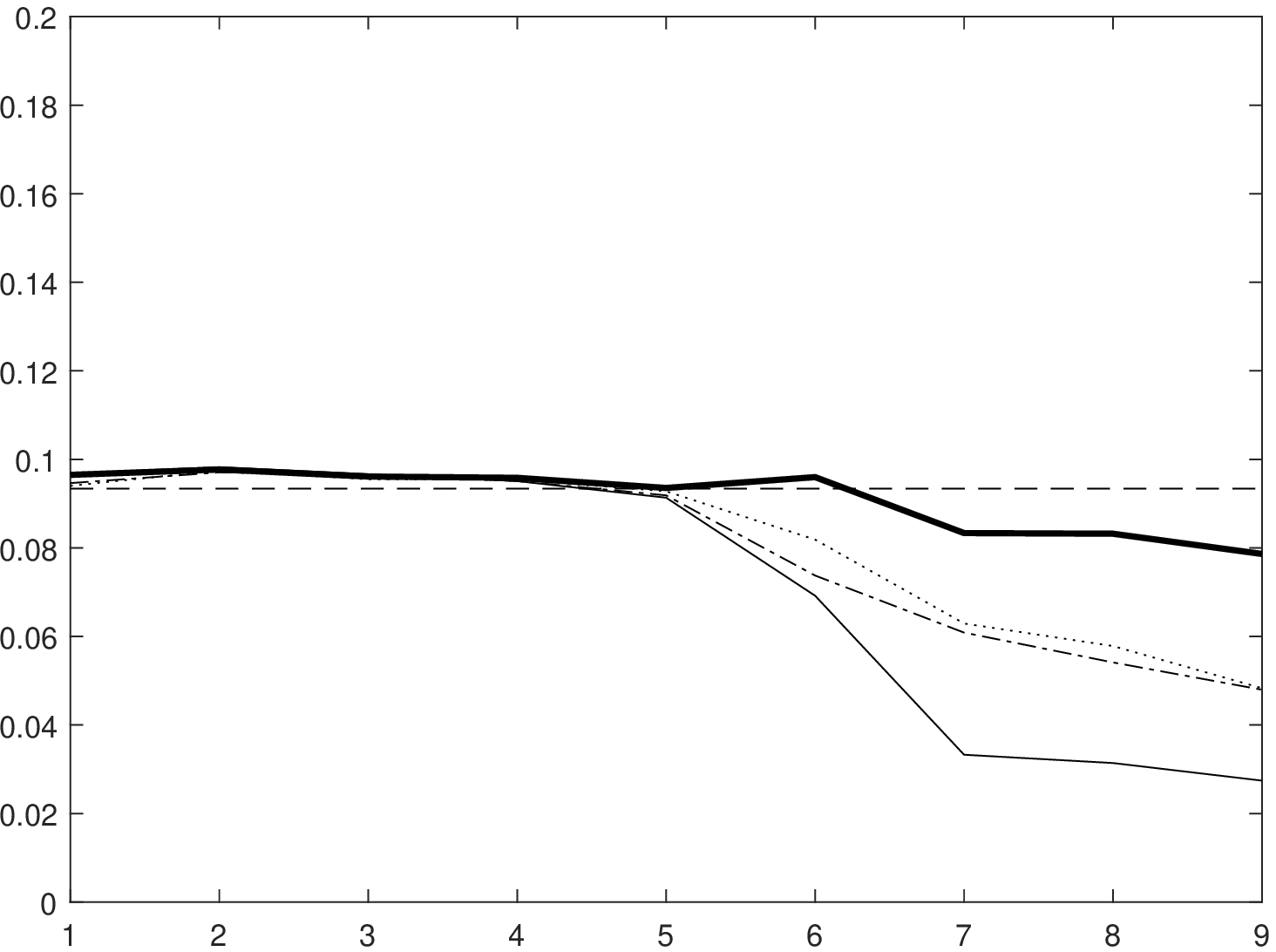}}
		\subfigure[Arm / QDA]{\label{f2}\includegraphics[width=3.75cm]{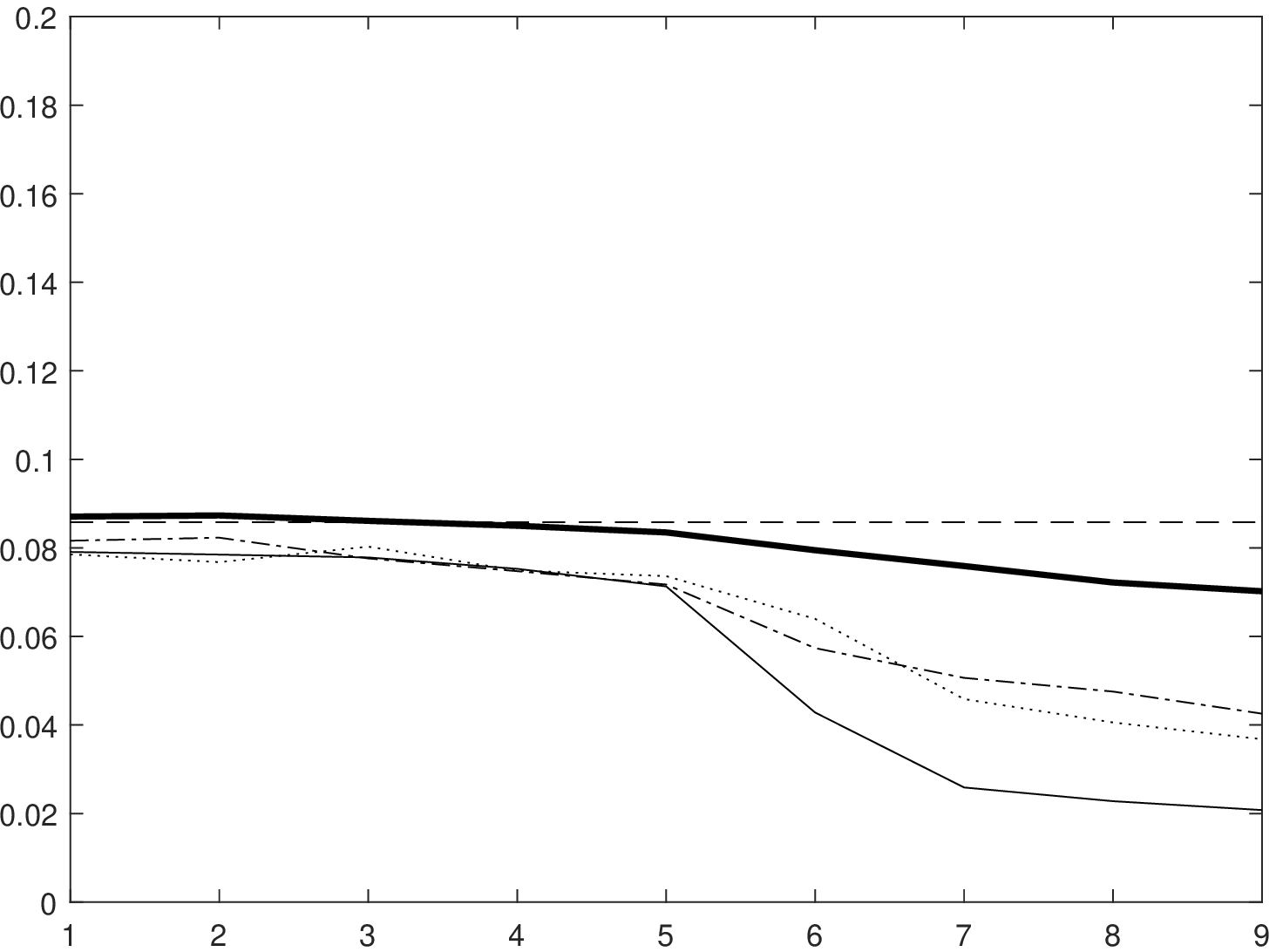}}
         \subfigure[Arm / CART]{\label{f3}\includegraphics[width=3.75cm]{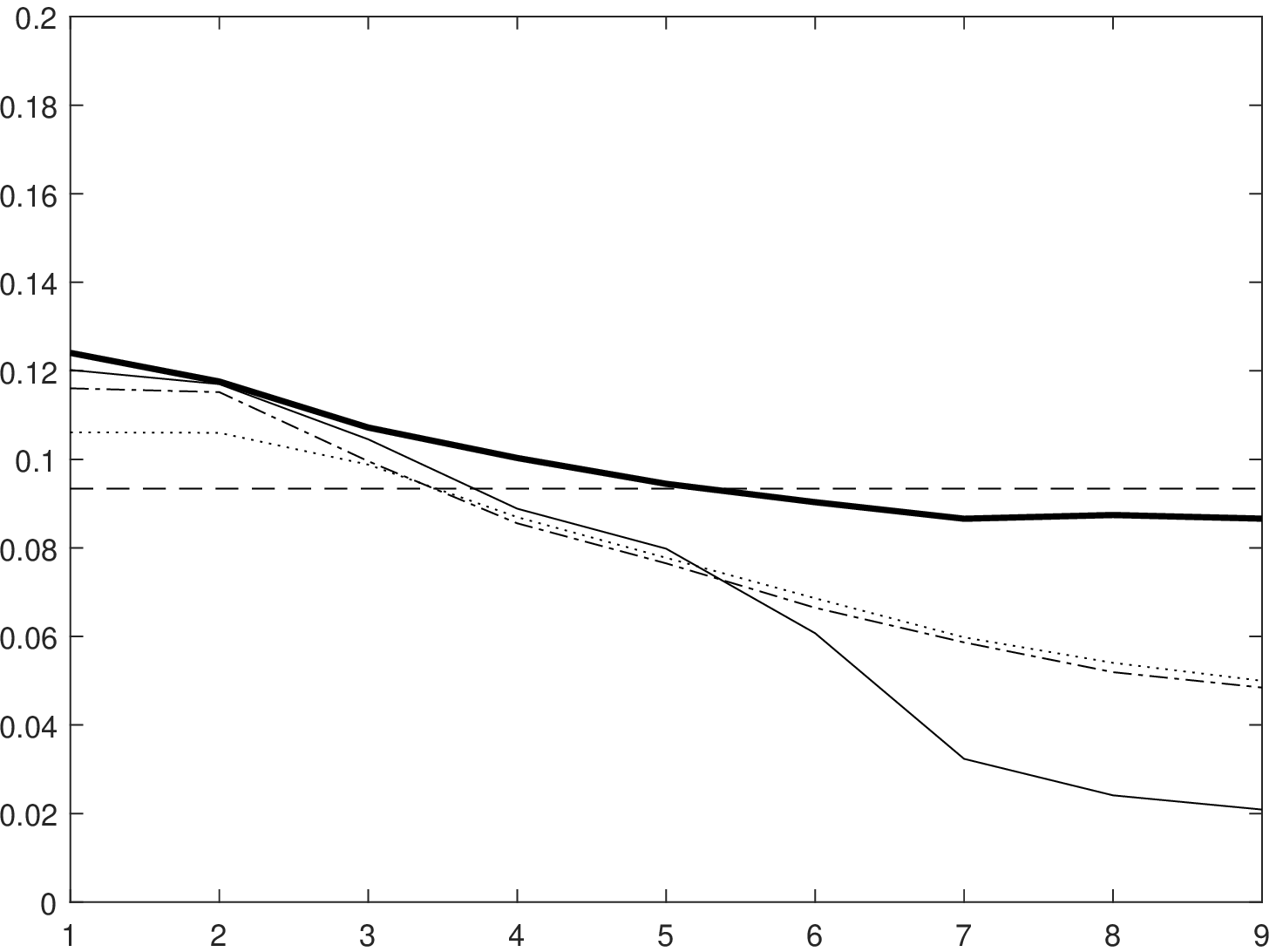}}
		\subfigure[Waist / LDA]{\label{f4}\includegraphics[width=3.75cm]{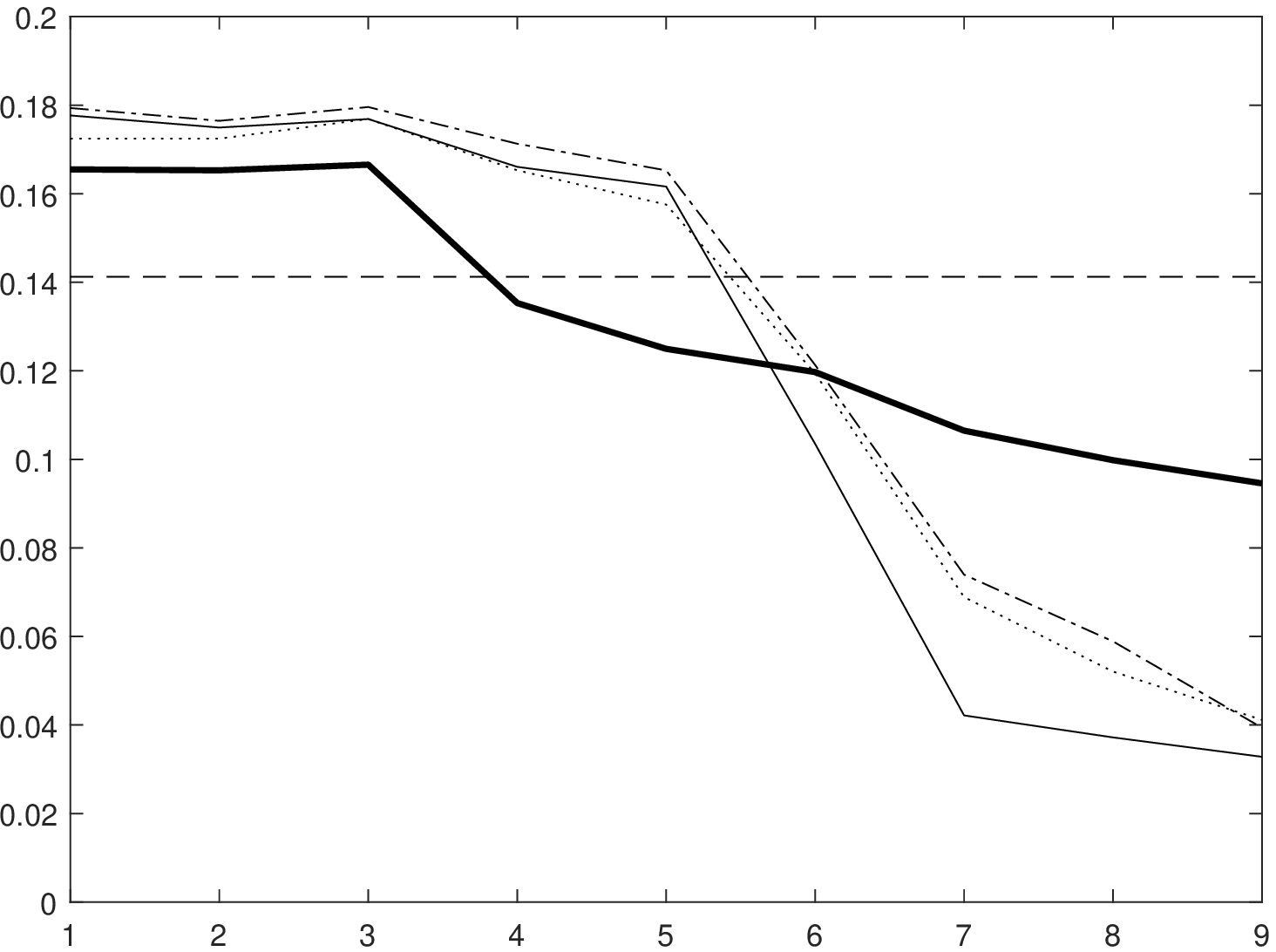}}
         \subfigure[Waist /  QDA]{\label{f5}\includegraphics[width=3.75cm]{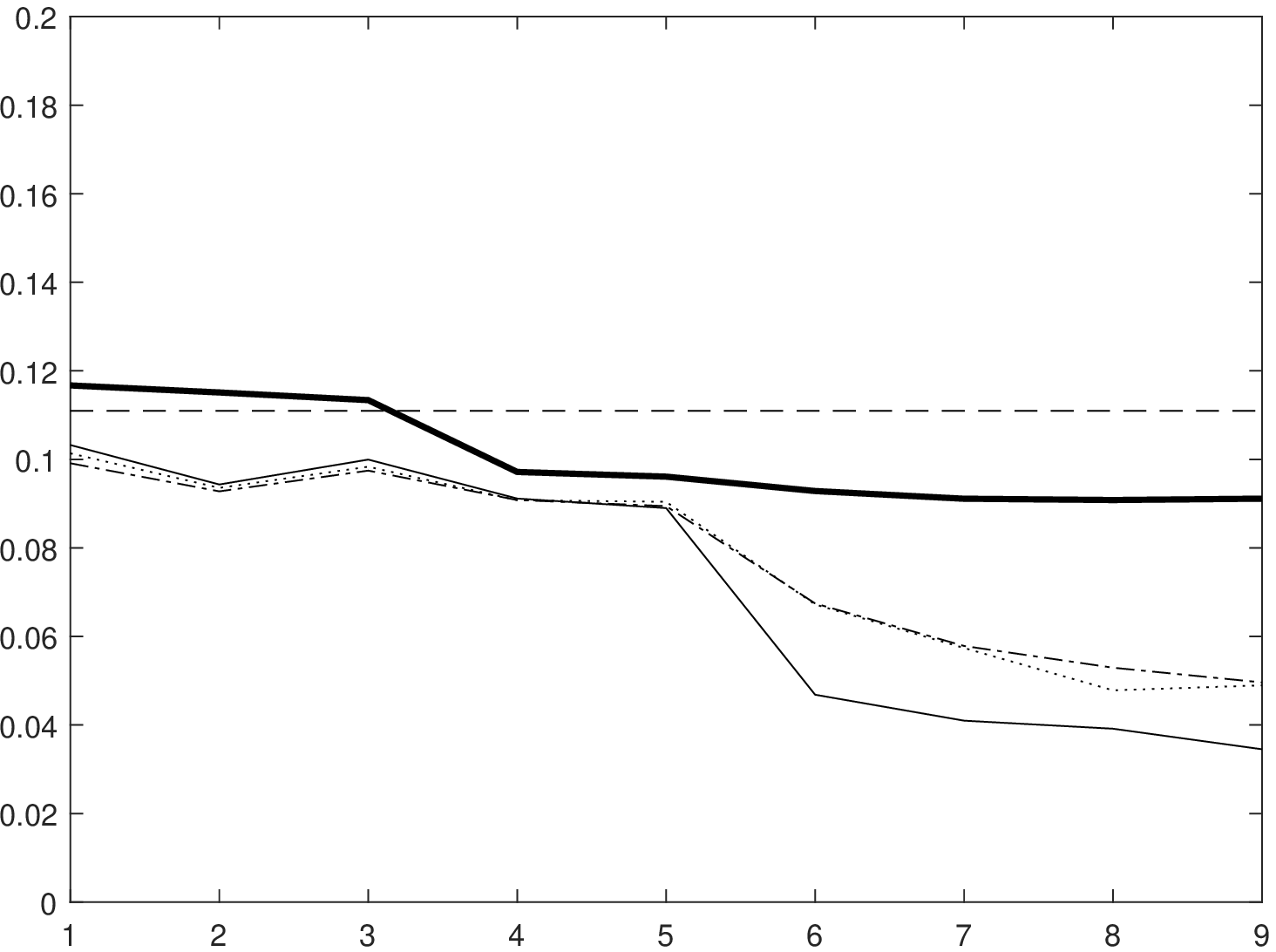}}
		\subfigure[Waist / CART]{\label{f6}\includegraphics[width=3.75cm]{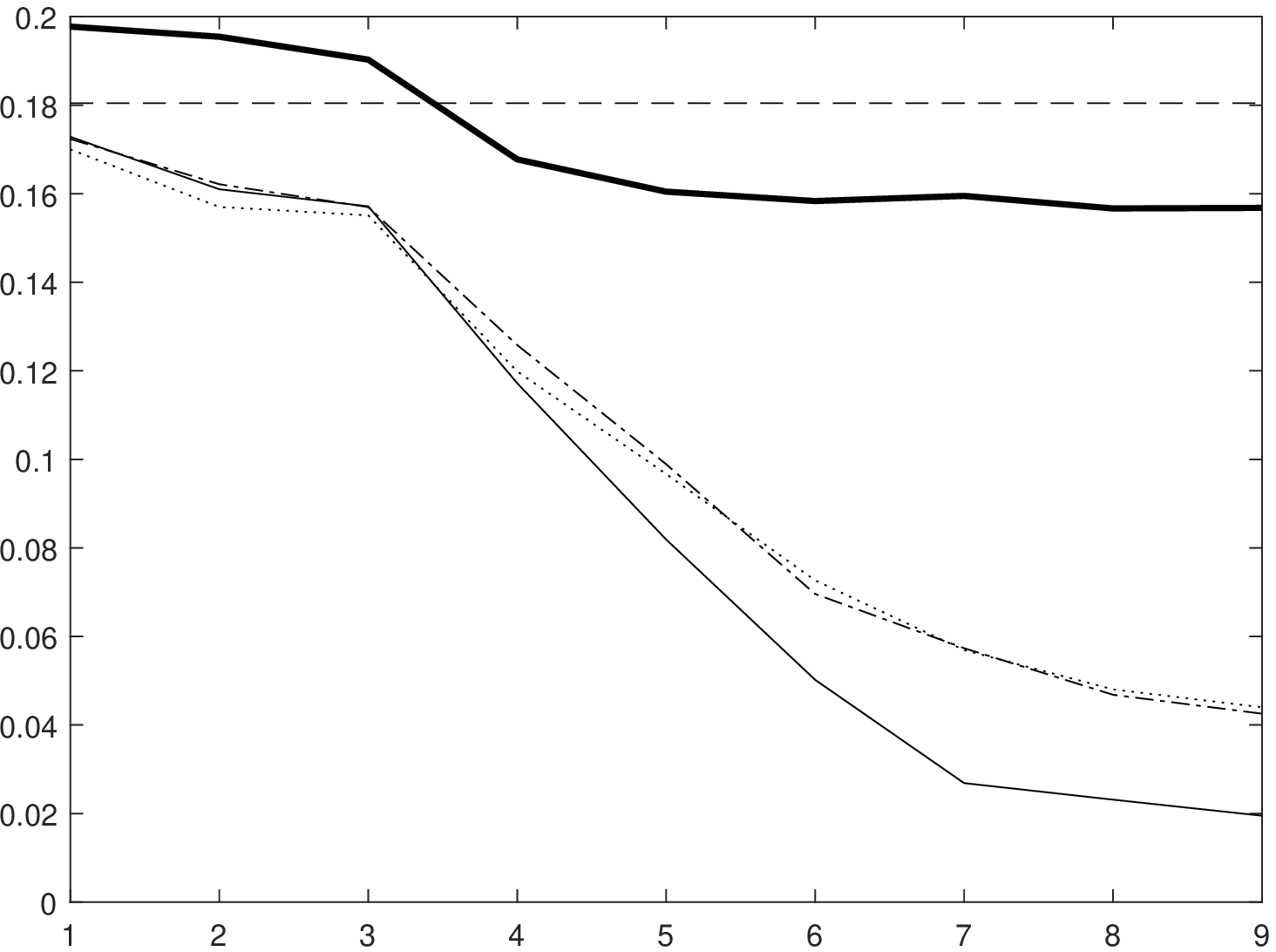}}
         \subfigure[Wrist / LDA]{\label{f7}\includegraphics[width=3.75cm]{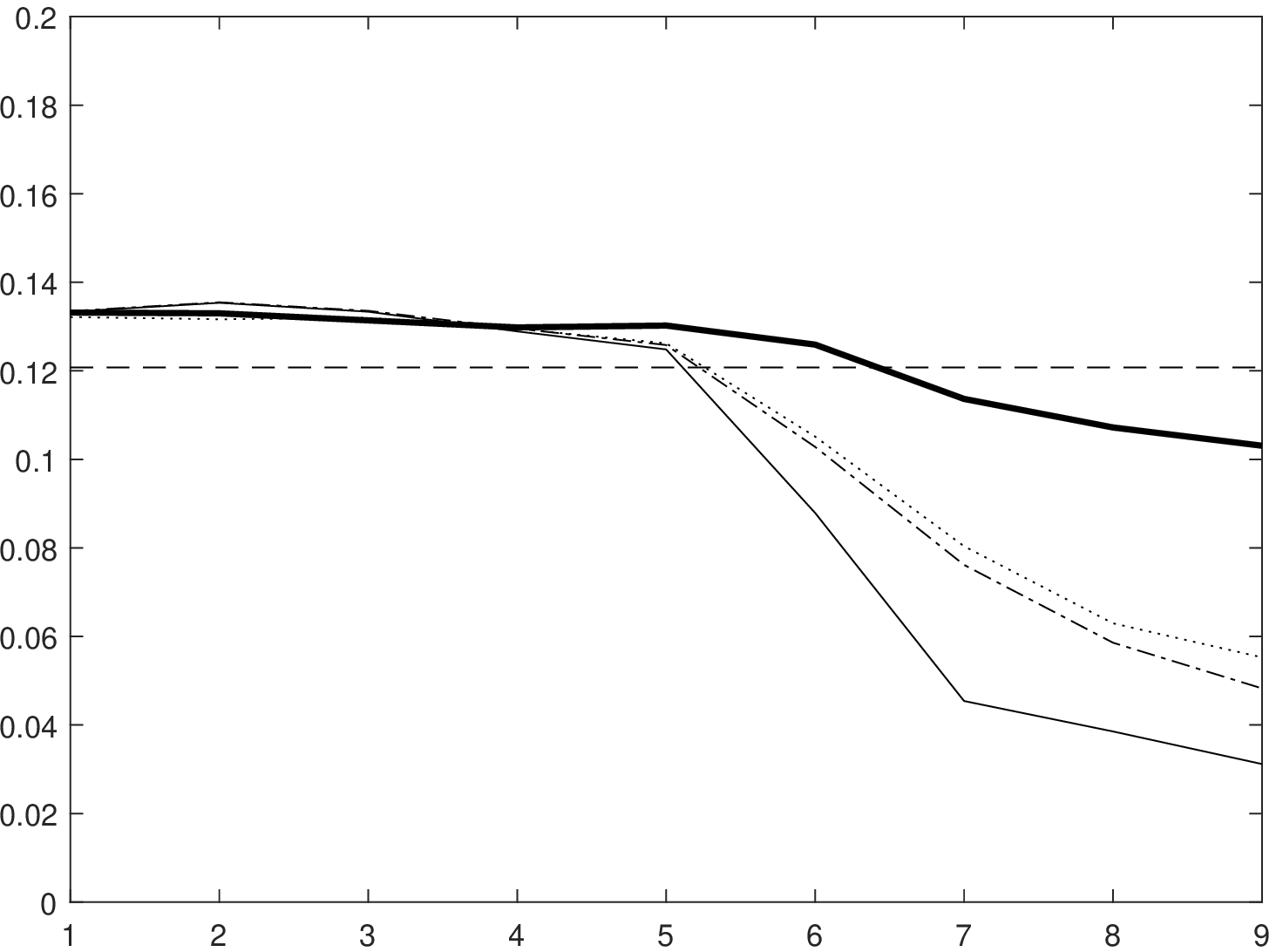}}
		\subfigure[Wrist/ QDA]{\label{f8}\includegraphics[width=3.75cm]{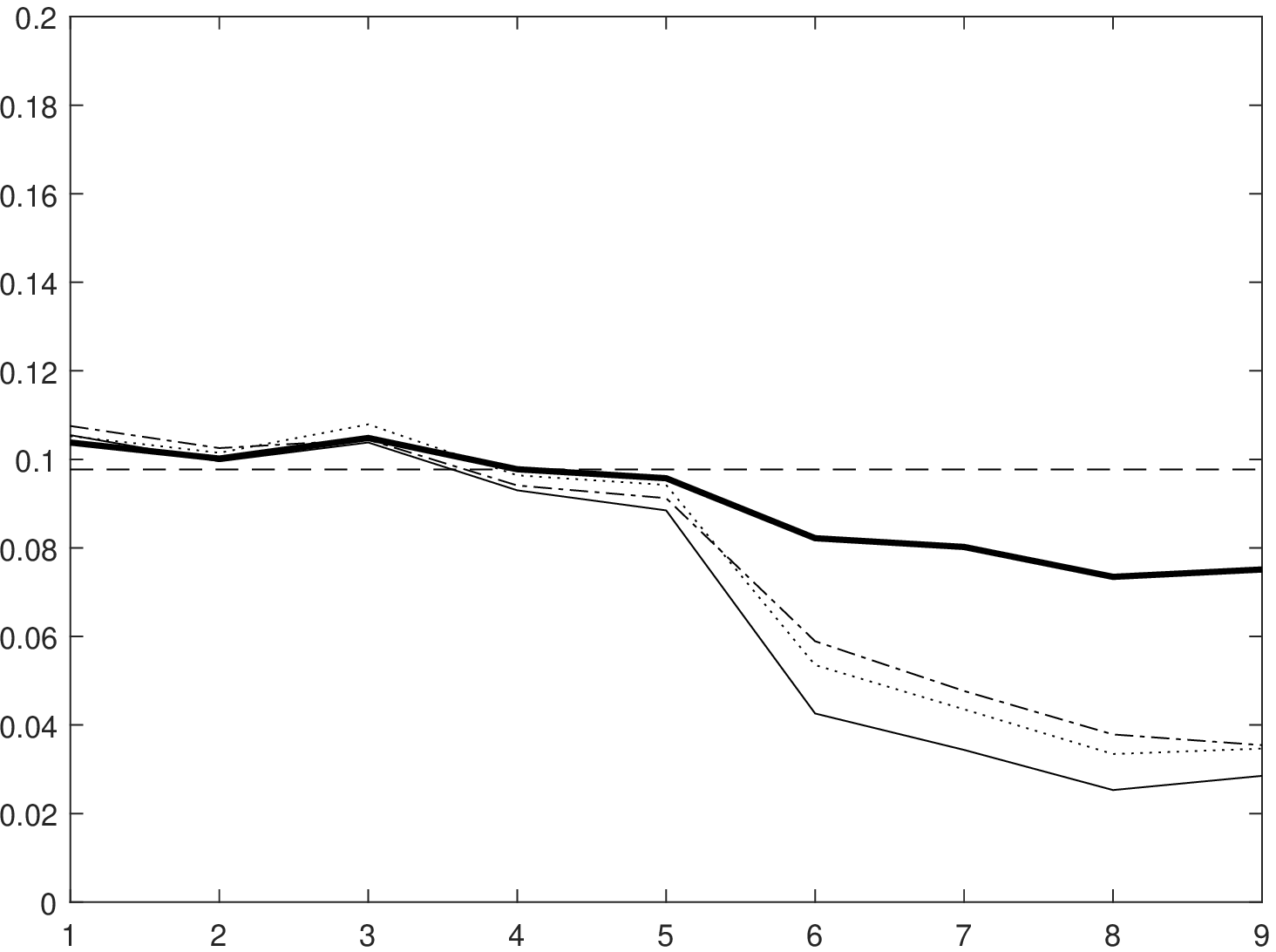}}
        \subfigure[Wrist / CART]{\label{f9}\includegraphics[width=3.75cm]{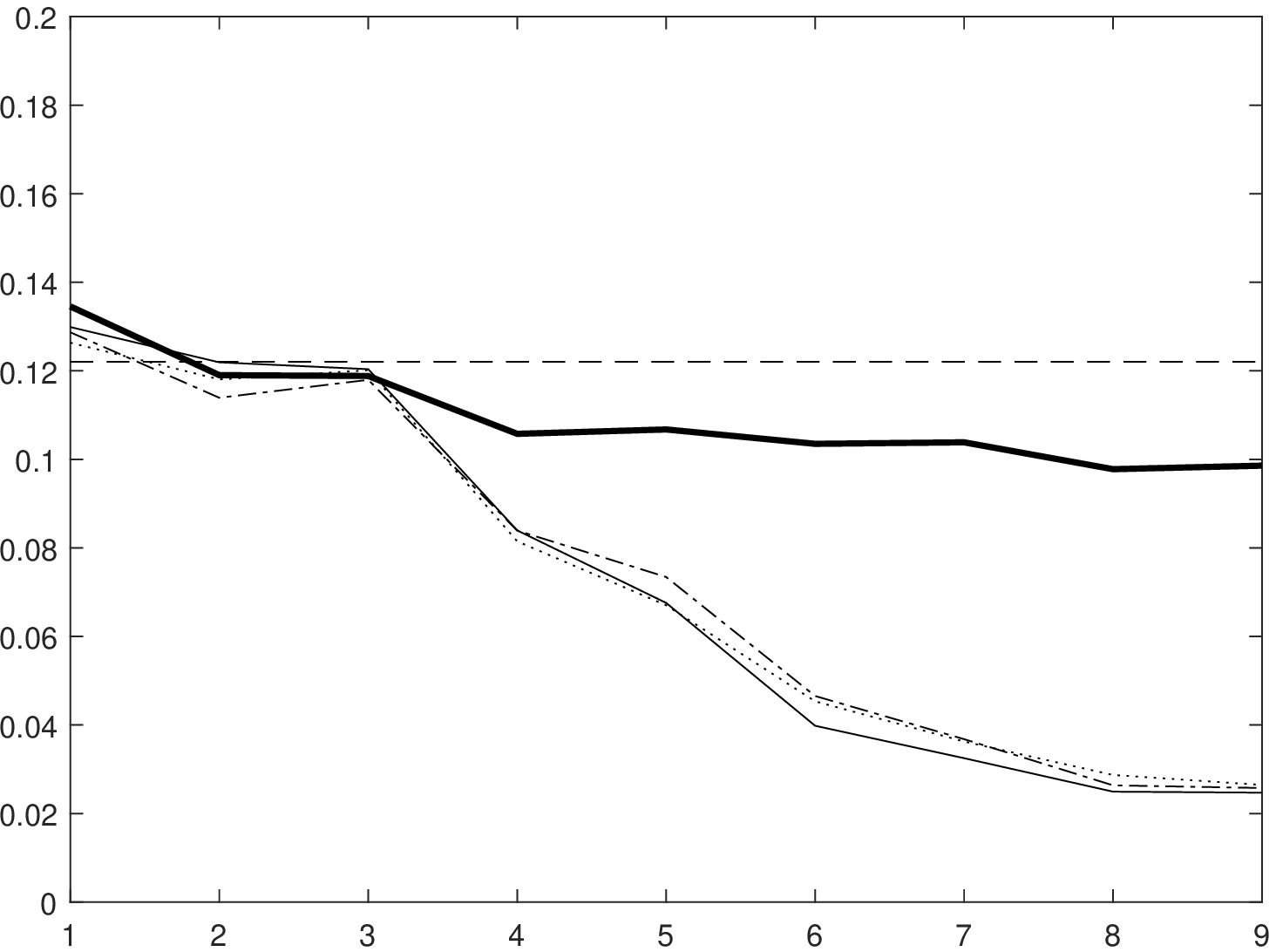}}
    \caption{Adding of new base models to Learn++ decreases the error rate. The error rate is shown in $y$-axis and $x$-axis shows the number of used base models. The first row shows results when the sensor position is arm, second waist, and third wrist. The first column shows results using LDA, second QDA, and third CART. User-independent results are shown using horizontal line, non-supervised using a thick solid line, supervised with a thin solid line, semi-supervised with $th=0.90$ using dotted line, and with $th=0.95$ using dash-dot line.} 
\label{inc_learn}
\end{center}
\vspace{-2em}
\end{figure}

\section{Experiments and discussion}
\label{experiments}
\vspace{-0.5em}

\begin{table}[t]
\vspace{-2em}

\begin{scriptsize}
\caption{Average error rates of different personalization approaches. The percentage of user inputs required by the semi-supervised approach in parentheses.}

\begin{center}
\vspace{-1em}
\begin{tabular}{lccccc}
     & User-indep. & Nonsuper & Semi-super, th=90& Semi-super, th=95& Superv.\\

    Arm / LDA & 9.3 &7.9 & 4.8 (9.9)&4.8 (10.1)& 2.7\\
    Waist / LDA & 14.1 & 9.5 & 4.1 (14.9)& 3.9 (14.9)& 3.3\\
    Wrist / LDA & 12.7 &10.3 & 5.5 (10.9)& 4.8 (11.8)& 3.1\\
    
    Arm / QDA & 8.6 &7.0 & 3.7 (13.0)& 4.3 (12.3)& 2.1\\
    Waist / QDA & 11.1 &9.1 & 4.9 (10.7)&5.0 (10.3)&3.5 \\
    Wrist / QDA & 12.2 &7.5 &3.5 (14.9) & 3.5 (15.7)& 2.9\\
    
    Arm / CART & 11.7 &8.7 &5.0 (23.0)&4.9 (22.8)& 2.1\\
    Waist / CART & 18.0 &15.7 &4.4 (27.4)&4.3 (28.6) & 2.0\\
    Wrist / CART & 12.2 &9.9 &2.6 (26.7)& 2.6 (27.0)& 2.5\\
    
    Mean & 12.2 &9.5 & 4.3 (16.8)& 4.2 (17.1)& 2.7 \\
\end{tabular}

\label{errrates}

\end{center}
\end{scriptsize}
\vspace{-2em}
\end{table}

For the experiments, each person's $s_i$ data are divided into three parts and each part contains the same amount of data from each activity \cite{siirtola2018ESANN}. When a user-independent model is personalized for subject $s_i$, two parts of the data are used for personalizing the model (in Steps 2 \& 3) and the last part is used for testing.

Non-supervised, semi-supervised (with threshold values 0.95 and 0.90), and supervised approaches to personalize recognition model were experimented. These are compared to user-independent recognition rates. Figure \ref{inc_learn} and Table \ref{errrates} show error rates from the balanced accuracies averaged over all nine study subjects. The benefit of using a  personalized model is obvious: in each case, the average error rate using a personalized model is lower than using a user-independent model no matter which personalization approach, body position, or base classifier is studied. In some cases, the difference is more than 10\%-units. 

Model update clearly benefits from user inputs as the error rates using semi-supervised approach (4.2/4.3\%) are much lower than non-supervised (9.5\%), no matter which base classifier is used. 
The difference is especially big in cases where the user-independent model, originally used in the recognition process, is not that accurate, see for instance Figures \ref{f4} \& \ref{f6}. In these cases the error rate of the semi-supervised approach starts to rapidly drop when personal base classifiers are added to the ensemble, while in the non-supervised approach, it does not drop that much. Eventually, the difference in error rates of non-supervised and semi-supervised ($th=0.95$) approaches is 5.7 and 11.4\%-unit, respectively. In these cases, a user-independent model cannot detect some of the classes at all, and therefore, the non-supervised model suffers from a drastic concept drift, what is more, it cannot recognize all the activities. However, the semi-supervised model can recover from such a situation because of user inputs. However, when supervised and semi-supervised results are compared, it can be noted that already by replacing a small number of instances with correct labels, almost as high recognition rates can be achieved as by labeling all the instances. On average, the error rate using the semi-supervised approach is less than 2\%-units higher than using the supervised approach.  Moreover, when semi-supervised approach was used, there does not seem to be differences between threshold values $th=0.95$ and $th=0.90$. This is because the used classifiers tend to give 100\% as a priori value or some much lower value. In fact, in both cases, approximately same amount of windows were labeled by the user (26\% using CART, 13\% using QDA, and 12\% using LDA, see Table \ref{errrates}). However, as also two windows before and after uncertain observation was labeled according to this same input, the amount of replaced labels was bigger (47\% using CART, 27\% using QDA, and 25\% using LDA). It can be noted that a priori values are highly dependent on the used classifier, and therefore, the threshold to conclude if the predicted classification is reliable should be selected classifier-wise.

\section{Conclusions}
\label{conclusion}
\vspace{-0.5em}

In this study, Learn++ was applied to human activity recognition data to personalize recognition models, and three different approaches to personalize modes were compared: non-supervised, semi-supervised, and supervised. Three base classifiers were compared: LDA, QDA and CART. Experiments showed that personalized models are more accurate than user-independent models. Moreover, it was noted that without user inputs, personalized models can suffer from a drastic concept drift, as the error rates of non-supervised models were much higher than the one's obtained using the semi-supervised approach (semi-supervised  4.2/4.4\% vs. non-supervised 9.5\%). However, using the proposed method, already by asking the user to label part of the data (from 12\% to 26\% depending on the used base classifier), almost as high recognition rates can be achieved as by using only user labeled data. In fact, the average error rate using the semi-supervised approach instead of supervised was less than 2\%-units. The future work includes experimenting with more extensive data sets.

\begin{scriptsize}

\end{scriptsize}

\end{document}